\documentclass{elsart}
\usepackage{amsfonts}
\usepackage{amsmath}
\usepackage{natbib}
\usepackage{url}
\usepackage{fancyhdr}

\newcommand{\R}[1]{\ensuremath{\Rset^{#1}}}

\newcommand{\N}[1]{\ensuremath{\Nset^{#1}}}

\usepackage[bookmarks=true,ps2pdf=true,hyperindex=true,colorlinks=false,
pdftitle={Functional Multi-Layer Perceptron: a Nonlinear Tool for Functional Data Analysis},
pdfkeywords={Functional data analysis, Multi-Layer Perceptron, Universal approximation, Supervised learning, Curves discrimination, Learning consistancy, Nonlinear functional model, Spectrometric data},
pdfsubject={Functional Data Analysis with Multi-Layer Perceptrons},
pdfauthor={Fabrice Rossi and Brieuc Conan-Guez}]{hyperref} 
\begin{document}
\date{22 July 2004}
\begin{frontmatter}
\title{Functional Multi-Layer Perceptron: a Nonlinear Tool for Functional Data
  Analysis\protect\footnote{Published in Neural Networks (Volume 18, Number 1,
  pages 45--60). DOI: \url{http://dx.doi.org/10.1016/j.neunet.2004.07.001}}}
\author[LISE,INRIA]{Fabrice Rossi\corauthref{rossi}}
\ead{Fabrice.Rossi@inria.fr}
and
\author[INRIA]{Brieuc Conan-Guez}
\ead{Brieuc.Conan-Guez@inria.fr}
\address[LISE]{CEREMADE, UMR CNRS 7534, Universit\'e Paris-IX Dauphine,
Place du Mar\'echal de Lattre de Tassigny, 75016 Paris, France}
\address[INRIA]{INRIAbi, Domaine de Voluceau, Rocquencourt, B.P. 105,
78153 Le Chesnay Cedex, France}
\corauth[rossi]{Corresponding author:\\
Fabrice Rossi\\
Projet AxIS\\
INRIA Rocquencourt\\
Domaine de Voluceau, Rocquencourt, B.P. 105\\
78153 LE CHESNAY CEDEX -- FRANCE\\
Tel: (33) 1 39 63 54 45\\
Fax: (33) 1 39 63 58 92\\
}
\end{frontmatter}

\newpage

\begin{frontmatter}
\title{Functional Multi-Layer Perceptron: a Nonlinear Tool for Functional Data
  Analysis}
\begin{abstract}
In this paper, we study a natural extension of Multi-Layer Perceptrons (MLP)
to functional inputs. We show that fundamental results for classical MLP can
be extended to functional MLP. We obtain universal approximation results that
show the expressive power of functional MLP is comparable to that of
numerical MLP. We obtain consistency results which imply that the estimation
of optimal parameters for functional MLP is statistically well defined. We
finally show on simulated and real world data that the proposed model performs
in a very satisfactory way. 
\end{abstract}
\end{frontmatter}

\newpage

\section{Introduction}
Functional Data Analysis (FDA, see \citet{RamseySilverman97} for a
comprehensive introduction to FDA methods) is an extension of traditional data
analysis to functional data. In this framework, each individual is
characterized by one or more real valued functions, rather than by a vector of
\R{n}. An important feature of FDA is its ability to take into account
dependencies between numerical measurements that describe an individual,
especially smoothness. If we represent for instance the size of a child at
different ages by a vector, traditional methods generally consider each value
to be independent of the others. In FDA, the size is represented as a
function (in general a regular one) that maps measurement times to
centimeters. 

In order to deal with irregular measurements and to allow numerical
manipulation of functions, FDA replaces actual observations by a simple
functional representation. Spline based approximation is the most commonly
used method, as it represents each individual by a smooth function. Kernel or
wavelet based approximations are also used. FDA has been successfully applied 
to real problems such as climatic variation forecasting
(\citet{BesseCardotStephenson00}), acidification process studying
(\citet{Abraham2000}), analysis of children size evolution
(\citet{RamseySilverman97}), land usage prediction based on satellite
  images (\citet{BesseEtAl2004FDA}), etc.

In this paper, we focus on a precise yet very general task: we assume that we
observe functions associated to a classical target variable. This variable can
be for instance a class label, in which case we perform supervised
classification. If the variable is a real valued vector, we perform a
regression. The key idea is that, whereas individuals are described thanks to
functions, we still want to predict a traditional numerical value. In
mathematical terms, we have $n$ examples described by $s+1$ variables,
$(g_1^i,\ldots,g_s^i,t^i)_{i\in \{1,\ldots,n\}}$, where $t^i$ is the target
variable (with $t^i\in\R{o}$) and where each $g^i_l$ is a function belonging
to a given functional space. The problem is to predict $t^i$ based on
$(g_1^i,\ldots,g_s^i)$.  In the framework of FDA, several methods have been
proposed to solve this kind of problem, for instance the linear functional
model (see e.g.  \citet{HastieMallows1993}, \citet{MarxEilers1996},
\citet{RamseySilverman97}, \citet{CardotFerratySarda1999},
\citet{CardotFerratySarda2002} and \citet{James2002}), functional discriminant
analysis (e.g.  \citet{JamesHastie2001}), functional Slice Inverse Regression
(see \citet{Li1991} for the classical SIR and \citet{FerreYao2003SIR} for its
functional version) and non-parametric kernel based functional estimators (see
\citet{FerratyVieu2002CS}, \citet{FerratyVieu2002CSDA} and
\citet{FerratyGoiaVieu2002Test}). 

In this paper, we show how Multi-Layer Perceptrons (MLP) can be directly
applied to functional data, so as to provide nonlinear semi-parametric
function classification and regression. We introduce a major difference with
traditional FDA methods: our model works directly with the studied functions,
without using a simplified representation. This avoids restrictions on the
functional weight representation which can therefore be adapted to the
context. For instance, functional data with low dimensional input spaces can
be manipulated thanks to generalized linear models (such as splines), whereas
MLP are used for functions with high dimensional input spaces.

When functional data are perfectly known, the extension of MLP we propose is a
particular case of an extension proposed and studied from purely theoretical
point of view in \citet{Stinchcombe99}. In \citet{Stinchcombe99}, the author
shows that traditional universal approximation results for MLP can be extended
to (almost) arbitrary input spaces, including infinite dimensional vectorial
spaces. These results rely on the approximation of continuous linear forms
defined on the MLP input space. In our work, we show how to carry out this
kind of approximation in practice, for instance by using traditional MLP. We
show this way that functional MLP are universal approximators and therefore
that they can be used to model complex dependencies between a real valued
target variable and functional inputs. 

Moreover, we show that training a parametric functional MLP on a finite
number of function examples is statistically valid, as the optimal
parameters obtained thanks to those examples provide a consistent estimation
of asymptotic optimal parameters, even if we assume limited knowledge on each
function example (i.e., each function is only known thanks to a finite number
of (input, output) pairs). This is a direct translation of classical
results, presented in \citet{White89Review} for instance, available for
numerical MLP. 

The rest of the paper is organized as follows. In the first part, we assume
that we have perfect knowledge of manipulated functions: we start by
introducing in section \ref{sectionModel} the proposed functional MLP model.
Then we show in section \ref{sectionUniversalApproximators} how the results of
\citet{Stinchcombe99} can be adapted to functional MLP to show they are
universal approximators. In the second part, we take into account sampling:
consistency of functional MLP training is studied in section
\ref{sectionConsistency}. Section \ref{sectionAlternative} compares our
approach to alternative neural solutions, on a theoretical point of
view. Then, section \ref{sectionExperiments} gives some 
experimental results both on simulated and real world data. Proofs are
presented in section \ref{sectionProofs}.

\section{Functional Multi-Layer Perceptrons}\label{sectionModel}
\subsection{Functional data}\label{subsecFunctionalData}
As stated in the introduction, an observation is described by $s+1$ values,
$(g_1,\ldots,g_s,t)$, where each $g_l$ is a function (and $t\in\R{o}$). More
precisely, we assume that $\mu_l$ is a $\sigma$-finite positive Borel measure
defined on \R{u_l} and that $g_l$ belongs to $L^{p_l}(\mu_l)$.

\subsection{Functional neurons}\label{subsecFunctionalNeurons}
The extension of numerical neurons to functional inputs is straightforward.
Indeed a $n $ input MLP neuron is characterized by a fixed activation
function, $ T $, a function from $ \R{} $ to $ \R{} $, by a vector from $
\R{n} $ (the weight vector, $ w $) and by a real valued threshold, $ b $.
Given a vectorial input $ x\in \R{n}$, the output of the neuron is
$N(x)=T(w.x+b) $.

This formula is based on the linear form $x\mapsto w.x$. When
$x=(g_1,\ldots,g_s)\in L^{p_1}(\mu_1)\times \ldots\times L^{p_s}(\mu_s)$, a
linear form can be constructed thanks to integrals, for instance:
\begin{equation}\label{eq:LinearFunction}
(g_1,\ldots,g_s)\mapsto \sum_{l=1}^s\int f_lg_l\, \td\mu_l,
\end{equation}
where $(f_1,\ldots,f_s)$ are measurable functions chosen such that
$f_lg_l\in L^1(\mu_l)$. Using this linear form, we can define a functional
neuron:
\begin{defn}
  A functional neuron on $E=L^{p_1}(\mu_1)\times \ldots\times L^{p_s}(\mu_s)$
  is defined thanks to a fixed activation function $ T $ from $ \R{} $ to $
  \R{} $, weight functions $f_l$ (such that $f_lg_l\in L^1(\mu_l)$) and a real
  valued threshold, $ b $. It calculates
\begin{equation}\label{eq:NeuronFunctionWeight}
N(g_1,\ldots,g_s)=T\left( b+\sum_{l=1}^s\int f_lg_l\, \td\mu_l \right).
\end{equation}
\end{defn}
This functional neuron is a special case of general neurons proposed in
\citet{Sandberg1996,SandbergXu1996,Stinchcombe99}. The main drawback of this
model is that it uses functional weights rather than numerical ones. This
problem can be solved by using parametric representation of functions. More
precisely, we assume given $s$ functions $F_1,\ldots,F_s$ such that
(hypothesis $H_a$):
\begin{enumerate}
\item $W_l\subset \R{v_l}$
\item $F_l$ is a function from $W_l\times \R{u_l}$ to \R{}
\item for each $w_l\in W_l$, $F_l(w_l,.)\in L^{q_l}(\mu_l)$ where $q_l$ is the
  conjugate exponent associated to $p_l$
\end{enumerate}
For instance, $F_l$ can be implemented thanks to a numerical MLP (in this
case, $w_l$ is the weight vector of the MLP) or thanks to the first functions
of a topological basis of $L^{ q_l}(\mu_l)$ (in this case, we have
$F_l(w_l,x)=\sum_{i=1}^{v_l}w_{li}\psi_i(x)$, where $(\psi_i)_{i\in\N{}}$ is
the considered topological basis). 

We can now introduce the definition of a parametric functional neuron:
\begin{defn}
  A parametric functional neuron on $E=L^{p_1}(\mu_1)\times \ldots\times
  L^{p_s}(\mu_s)$ is defined thanks to a fixed activation function $ T $ from
  $ \R{} $ to $ \R{} $, a weight vector $w\in W_1\times\ldots\times W_s$ and a
  real valued threshold, $ b $. It calculates
\begin{equation}\label{eq:FunctionalNeuron}
N(g_1,\ldots,g_s)=T\left( b+\sum_{l=1}^s\int F_l(w_l,x)g_l(x)\, \td\mu_l(x)
\right).
\end{equation}
\end{defn}

\subsection{Functional MLP}\label{subsecFunctionalMLP}
As a functional neuron gives a real output, we have to use numerical neurons
except in the first layer of a functional MLP. In particular, a one hidden
layer parametric functional perceptron with one functional input and one real
output computes a function of the following form:
\begin{equation}\label{eqFMLP}
H(g)=\sum ^{k}_{i=1}a_{i}T\left( b_{i}+\int F_i(w_i,x)g(x)\, \td\mu(x)\right),
\end{equation}
where the $a_i$ are real numbers, as well as the $b_i$, and $w_i$ are parameter
vectors for $F_i$. 

Of course, it is obvious to extend those definitions to more than one output
and/or hidden layer. The only difference between a functional $n$-hidden
layer perceptron and a numerical one is that, as stated above, we use
functional neurons only in the first layer. It is also obvious to define a
general functional MLP by using functional neurons rather than parametric
functional neurons. 

\section{Universal approximation}\label{sectionUniversalApproximators}
\subsection{Definitions and notations}
We use notations and definitions from \citet{Stinchcombe99}.
\subsubsection{Functional spaces and metrics}
We denote $C(A,B)$ the set of continuous functions from $A$ to $B$, where $A$
and $B$ are two topological spaces. As a special case, $C^n$ is the set of
continuous functions from \R{n} to \R{}. $M^n$ is the set of (Borel)
measurable functions from \R{n} to \R{}. We denote $d_C$ the metric on $M^n$
that gives uniform convergence over compact subsets:
\begin{equation}
d_C(f,g)=\sum_{n\in\N{*}}\frac{1}{2^n}\min\left\{\sup_{|x|\leq
n}|f(x)-g(x)|,1\right\}.
\end{equation}
When $K$ is a compact subset of $X$ a topological space, we define $\rho_K$ a
metric on the set of functions from $K$ to \R{} by:
\begin{equation}
\rho_K(f,g)=\sup_{x\in K}|f(x)-g(x)|.
\end{equation}
\begin{defn}
Let $X$ be a metric space with $d$ the associated metric. Let $C$ and $S$ be
two subsets of $X$. $S$ is $d$-outside dense in C if the $d$-closure of $S$
contains $C$, and $S$ is $d$-inside dense in $C$ if the $d$-closure of $S\cap
C$ contains $C$. 
\end{defn}
When $C=X$, $d$-inside density is equivalent to $d$-outside density and is
simply called $d$-density. 

\subsubsection{One hidden layer perceptrons}
\begin{defn}
If $ T $ is a function from $ \R{}  $ to $ \R{}  $ and $ n $
a positive integer, $ S_{T}^{n} $ is the set of functions exactly
computed by one hidden layer perceptrons with $ n $ inputs and
one output, and using $ T $ as activation function, i.e. the set
of functions of the form $ h(x)=\sum _{i=1}^{p}\beta _{i}T(w_{i}.x+b_{i}) $
where $ p\in N $, $ \beta _{i}\in \R{}  $, and $ (w_{i},b_{i})\in \R{n+1} $. 
\end{defn}
\begin{defn}
If $ X $ is a topological vector space, $ A $ a subset of $ X^{*} $ and $ T $
a function from $ \R{} $ to $ \R{} $, $ S_{T}^{X}(A) $ is the set of functions
exactly computed by one hidden layer generalized perceptrons with input in
$X$, one real output, and weight forms in $A$, i.e. the set of functions from
$ X $ to $ \R{} $ of the form $ h(x)=\sum ^{p}_{i=1}\beta
_{i}T(l_{i}(x)+b_{i}) $ where $ p\in N $, $ \beta _{i}\in \R{} $, $ b_{i}\in
\R{} $ and $ l_{i}\in A $. 

Note that $A$ can in fact be any set of functions from $X$ to \R{}, in which
case  we do not introduce constant terms $b_i$.
\end{defn}
According to this definition, functional one hidden layer
perceptrons are a special case of Stinchcombe generalized perceptrons in which
$X$ is a product of $L^p$ spaces and $A$ is given by linear forms of the form
$l(g_1,\ldots,g_s)=\sum_{l=1}^s\int f_lg_l\, \td\mu_l$ (or
$l(g_1,\ldots,g_s)=\sum_{l=1}^s\int F_l(w_l,x)g_l(x)\, \td\mu_l(x)$ for
parametric functional perceptrons). 

\subsection{Universal approximation with functional MLP}
Several approximation results show that $ S_{T}^{X}(A) $ is inside or outside
dense in different functional spaces. Indeed \citet{Stinchcombe99} (as well as
\citet{SandbergXu1996} and \citet{Chen1998}) proposes approximation results
for $S_{T}^{X}(A)$ for almost arbitrary spaces $X$ (see theorem 5.1 and
corollaries 5.1.2 and 5.1.3 from \citet{Stinchcombe99}). In order to apply
those general results to practical cases, complex technical properties have to
be satisfied by $A$. In this section, we show that those properties are
satisfied by very general functional one hidden layer perceptrons.
\begin{cor}
\label{corollary:symbolicNN:universal}Let $ \mu  $ be a finite
positive Borel measure on $ \R{n} $.  Let $ 1<p\leq \infty $ be an arbitrary
real number and $ q $ be the conjugate exponent of $ p. $ Let $V$ be a dense
subset of $ L^{q}(\mu ) $. Let $ A_V $ be the set of linear forms on $
L^{p}(\mu ) $ of the form $ l(f)=\int fg\, \td\mu $, where $ g\in V $.  Let $ T
$ be a measurable function from $ \R{} $ to $ \R{} $ such that $S_T^1$ is
$d_C$-inside (resp. $d_C$-outside) dense in $C^1$. Then $ S_{T}^{L^{p}(\mu
  )}(A_V) $ is $\rho_K$-inside (resp. $\rho_K$-outside) dense in $ C(K,\R{} )
$, where $ K $ is any compact subset of $L^{p}(\mu ) $.
\end{cor}
\begin{cor}
\label{corollary:symbolicNN:lone}Let $ \mu  $ be a finite positive
compactly supported Borel measure on $ \R{n} $. Let $ T $ be a measurable
function from $ \R{} $ to $ \R{} $, such that $S_T^1$ is $d_C$-inside (resp.
$d_C$-outside) dense in $C^1$.  Let $V$ be a subset of $L^\infty(\mu)$
$d_C$-inside (or $d_C$-outside) dense in $C^n$. Then $ S_{T}^{L^{1}(\mu
  )}(A_V) $ is $\rho_K$-outside dense in $ C(K,\R{} )
$, where $ K $ is any compact subset of $ L^{1}(\mu ) $.
\end{cor}

\subsection{Discussion}
Corollary \ref{corollary:symbolicNN:universal} shows that as long as we can
approximate functions in $L^q(\mu)$ and in $C^1$, then an one hidden layer
perceptron can be used to approximate functions in $ C(K,\R{} ) $ , where $ K
$ is a compact subset of $ L^{p}(\mu ) $. Previous works give very weak
conditions on $T$ that imply $d_C$ inside or outside density of $S_T^1$ for
$C^1$, see for instance Theorem 1 in \citet{Leshno93} and Theorem 1 in
\citet{Hornik93}. Basically, $T$ must be non polynomial and Riemann integrable
on a non-degenerate compact interval of \R{}, properties that are obviously
satisfied by popular activation functions such as $\tanh$. 

The generalized MLP used in corollary \ref{corollary:symbolicNN:universal}
uses linear forms in $A_V$ and is therefore a functional MLP with weight
functions chosen in $V$ a dense subset of $L^q(\mu)$. In practical situation,
weight functions are represented thanks to parametric functions
($F(w,.)$). This constraint does not introduce any problem, as long as we
choose a parametric universal approximator for $L^q(\mu)$. Thanks to Theorem 1
of \citet{Hornik91}, we can use for instance one hidden layer perceptrons
based on activation function $U$ (i.e., $V=S_{U}^n$) as long as $U$ is
measurable, bounded and non constant (as $p>1$, $q<\infty$ and Theorem 1
applies). Other models can be used (B-spline, wavelet, Fourier series, etc.)
but imply in general additional restrictions on the considered functional
space.

The proof of corollary \ref{corollary:symbolicNN:universal} could be extended
to $p=1$, and therefore, one might wonder why corollary
\ref{corollary:symbolicNN:lone} is useful. As pointed out in the introduction
of \citet{Stinchcombe99}, no $S_T^n$ set is dense in $L^{\infty}(\mu ) $.
Therefore, corollary \ref{corollary:symbolicNN:universal} main assumption ($V$
is dense in $L^{q}(\mu )$) cannot be satisfied by MLP based approximation.
This reduces greatly the interest of corollary
\ref{corollary:symbolicNN:universal} for $p=1$. That's why corollary
\ref{corollary:symbolicNN:lone} is useful: as shown for instance by Theorem 1
of \citet{Hornik93}, $S_{U}^n$ can be used to provide approximation to
continuous functions on a compact set. Therefore, the situation for $p=1$ is
quite similar to the one that stands for $p>1$, except that the measure has to
be compactly supported.

This means that when $K$ is a compact subset of a $L^p(\mu)$ functional space,
any function from $ C(K,\R{} ) $ can be approximated to a given precision
level by a functional MLP that uses a finite number of parameters (because
linear forms can be represented for instance thanks to numerical MLPs).
Despite the radical change in the input space dimension (from \R{n} to a
compact subset of a functional space), we can still effectively approximate
continuous functions.

It is very common in FDA to assume that studied functions are smooth, that is
at least continuous. If we only consider compact input spaces for those
functions, their case is covered by corollary
\ref{corollary:symbolicNN:universal}. Indeed, continuous functions (or more
regular functions) on a compact subset $Z$ of $\R{n}$ are obviously elements
of $L^\infty(\lambda)$ where $\lambda$ is the restriction of the Lebesgue
measure to $Z$. Moreover a compact subset $K$ of a space of regular
functions (considered with the uniform norm) is a compact subset of
$L^\infty(\lambda)$. This means that any continuous function from $K$ to \R{}
can be approximated by a functional MLP as long as $L^1(\lambda)$ can also be
approximated (this can be done thanks to $S_{U}^n$ \citet{Hornik91}). 

Extension of proposed corollaries to multiple functional inputs is
straightforward. In fact, corollaries are based on approximation of linear
forms on $X$ the input space of extended neurons. When
$X=L^{p_1}(\mu_1)\times\ldots\times L^{p_r}(\mu_r)$, approximation of elements
of $X^*$ is obtained thanks to approximations of elements of
$(L^{p_i}(\mu_i))^*$, because a linear form on $X$ is a linear combination of
linear forms on $L^{p_i}(\mu_i)$ (this fact was used to define the functional
neuron). 

\section{Consistency of Functional MLP learning}\label{sectionConsistency}
\subsection{Introduction}
As explained in the introduction, our goal is to explain a target variable
$t\in \R{o}$ thanks to functional observations $(g_1,\ldots,g_s)$. Basically,
we assume that there is a functional relationship such that $t\simeq
F(g_1,\ldots,g_s)$ and we try to model $F$ thanks to a functional MLP. Thanks
to universal approximation results given in the previous section, we know that
any regular $F$ can be approximated by a functional MLP. Nevertheless, an
important problem remains: $F$ is obviously unknown and a correct approximation
as to be constructed thanks to a limited number of examples of this
mapping. 

\subsection{Probabilistic framework}\label{subsectionProbaFramework}
\subsubsection{Functional data}
Let us now describe the probabilistic framework of our problem. All random
quantities will be defined on a given probability space $ (\Omega
,\mathcal{A},P) $. For the sake of simplicity, we consider only the case of an
unique functional input. More precisely, we make the following hypothesis
($H_b$):
\begin{enumerate}
\item $Z$ is a compact subset of $\R{u}$ 
\item $(G^i,T^i)_{i\in\N{}}$ is an i.i.d. sequence of random elements with
  values in $C(Z,\R{})\times \R{o}$ (i.e., each $G^i$ is a measurable function
  from $\Omega$ to   $C(Z,\R{})$ considered with its Borel sigma algebra and
  each $T^i$ is a random vector in $\R{o}$, and the sequence is i.i.d.)
\end{enumerate}
Hypothesis on the observed functions are quite different from those of
corollaries \ref{corollary:symbolicNN:universal} and
\ref{corollary:symbolicNN:lone}: on the one hand $H_b$ are stronger than
corollaries hypothesis as they consider only continuous functions defined on a
compact set, on the other hand they are weaker as observed functions do not
belong to a compact subset of $C(Z,\R{})$.

\subsubsection{Parametric model}
We try to model the relationship between $G^i$ and $T^i$ thanks to a special
kind of parametric model (a parametric functional MLP) that has the following
form: 
\begin{equation}\label{eq:ParametricModel}
H(w,g)=U\left(w_0,\int F_1\left( w_1,x\right) g(x)\td\mu(x),\ldots,
\int F_k\left( w_k,x\right) g(x)\td\mu(x)\right),
\end{equation}
where $w=(w_0,w_1,\ldots,w_k)\in W=W_0\times W_1\times \ldots\times W_k$, the
$F_l$ are parametric models as in parametric neurons, $U$ is a regular
function and $\mu$ a finite positive Borel measure (defined on $Z$). This
parametric form is quite similar to the one proposed in the context of Slice
Inverse Regression by \citet{FerreYao2003SIR}. Our main motivation here is to use
a general form that covers functional multi-layer perceptrons without making
too much hypothesis on their architecture (number of layers, activation
functions, linear terms, etc.). For instance, if $U$ is defined as follows:
\begin{equation}\label{eq:UForSHLP}
U(w_0,o_1,\ldots,o_k)=\sum_{l=1}^ka_lT(b_l+o_l),
\end{equation}
with $w_0=(a_1,b_1,\ldots,a_k,b_k)$, then $H(w,g)$ is exactly the output of a
functional one hidden layer perceptron, as given by equation \ref{eqFMLP}. As
a side effect, we cover any model that uses integrals to transform an input 
function into a real number. 

Some restrictions are needed on $F_l$ functions and on $U$ (hypothesis $H_c$):
\begin{enumerate}
\item for $0\leq l\leq k$, $W_l$ is a compact subset of \R{v_l}
\item for $1\leq l\leq k$, $F_l$ is a function from $W_l\times Z$ to \R{} such
  that:
  \begin{enumerate}
  \item for each $x\in Z$, $F_l(.,x)$ is continuous
  \item for each $w_l\in W_l$, $F_l(w_l,.)$ is measurable 
  \item $F_l$ is dominated on $W_l$, i.e., there is a measurable function
    $d_l\in L^p(\mu)$ (with $p\geq 1$) such that for for all $w\in W_l$ and
    $x\in Z$, $|F_l(w,x)|\leq d_l(x)$.
  \end{enumerate}
\item $U$ is an uniformly continuous function from $W_0\times\R{k}$ to \R{o}
\item $U$ is bounded
\end{enumerate}
Hypothesis $H_c$ are quite natural and are fulfilled in practical settings:
\begin{itemize}
\item Compacity of the parameter space is a classical hypothesis in
  consistency results.
\item Useful choices for $F_l$ are numerical MLP and basis expansions: for the
  former, continuity is mandatory in practice as optimal parameters are
  obtained thanks to gradient based algorithms (and therefore $F_l$ is in
  general differentiable with respect to $w_l$); for the latter, continuity is
  obvious as $F_l$ is linear with respect to $w_l$.
\item As stated before, when $F_l$ is obtained thanks to a numerical MLP, it
  is a continuous function. As $W_l$ and $Z$ are compact sets, the domination
  hypothesis is automatically fulfilled. When $F_l$ is obtained thanks to
  basis expansion, a natural hypothesis is to assume that basis functions
  belong to $L^p(\mu)$. Then, compacity of $W_l$ implies again that the
  domination hypothesis is fulfilled.
\item $U$ corresponds to the non functional part of a functional MLP, it is
  in general natural to assume that it is uniformly continuous. Indeed,
  popular activation functions such as $\tanh$ and the logistic function are
  uniformly continuous and moreover, $W_0$ is compact, therefore when $U$
  represents a MLP based on standard activation functions, it is uniformly
  continuous. Moreover, popular activation functions are also bounded and the
  assumption that $U$ is bounded is also natural. 
\end{itemize}

\subsubsection{Optimal model and consistency}
The learning phase in neural network applications consists in finding the best
parameters for a given task. In our framework, we assume given a
distance\footnote{$c$ has not really to be a distance, it can be any continuous
  positive function.} $c$
on \R{o} and we assess the quality of the neural model at the evaluation point
$G^i$ thanks to $c(T^i,H(G^i,w))$. We define the global error made by
the parametric model $H$ for parameters $w\in W$ by:
\begin{equation}\label{eq:TheoreticalError}
\lambda(w)=E\left(c(T^1,H(G^1,w))\right),
\end{equation}
where $E$ means expectation. Learning is in fact a parameter estimation
problem in which we try to optimize $\lambda(w)$ in order to find a
vector $w\in W^*$, where $W^*\subset W$ is the set of minimizer of
$\lambda(w)$. The practical problem is that $\lambda(w)$ cannot be exactly
calculated and is approximated thanks to a finite number of realizations of
$(G^i,T^i)$. More precisely, we define an empirical error by:
\begin{equation}\label{eq:EmpiricalError}
\widehat{\lambda}_n(w)=\frac{1}{n}\sum_{i=1}^nc(T^i,H(G^i,w)).
\end{equation}
This empirical error can be minimized to produce $\widehat{w}_n$ an estimation
of an optimal parameter vector. \citet{White89Review} shows that for numerical
MLP, $\widehat{w}_n$ is a strongly consistent estimation of an optimal
parameter vector. More precisely, if $d$ denotes the distance on $W$, then
$\lim_{n\rightarrow\infty}d(\widehat{w}_n,W^*)=0$ almost surely. Among
technical hypothesis needed to ensure this result, we adapt a domination
hypothesis to the functional framework (hypothesis $H_d$): $c(T^i,H(G^i,w))$
has to be dominated, in the sense that there is a positive function $c_{\max}$
from $\R{o}$ to \R{} such that:
\begin{enumerate}
\item $\forall w\in W,\ g\in C(Z,\R{})$ and $t\in \R{o}$, $c(t,H(g,w))\leq
  c_{\max}(t)$ 
\item $E(c_{\max}(T_1))<\infty$
\end{enumerate}
For functional MLP, hypothesis $H_d$ are quite natural. Indeed, hypothesis
$H_c$ (4) makes $H(g,w)$ bounded and therefore domination turns into an
hypothesis on $T_1$ and $c$. For instance if $c$ is the Euclidean distance in
$\R{o}$, then domination is obtained if $T_1$ has a second order moment.

Compared to the numerical case, we have two additional difficulties in the
functional framework: we are working with random elements with values in a
functional space, whereas \citet{White89Review} assumes that observations
belong to a finite dimensional space; moreover, perfect knowledge of observed
functions is seldom the case and we have to take into account that functions
are measured at a finite number of observation points.

\subsubsection{Function observations}\label{subsubsectionFunctionObservations}
In practical situations, each observed function is described by a finite
number of input/output pairs, such as $(x_j,g(x_j))_{j\in\{1,\ldots,m\}}$. We
choose the following mathematical model (hypothesis $H_e$):
\begin{enumerate}
\item $(X^{i}_{j})_{i\in\N{}\, j\in\N{}}$ is a sequence of independent
  sequences of random variables defined on $ (\Omega ,\mathcal{A},P) $ and
  with values in $Z$.
\item All $X^i_j$ are identically distributed and the induced probability
  measure on $Z$ is $\mu=P_X$. 
\item $(\mathcal{E}^{i}_{j})_{i\in\N{}\, j\in\N{}}$ is a sequence of
  independent sequences of random variables defined on $ (\Omega
,\mathcal{A},P) $ and with values in $\R{}$.
\item For all $i$, $(\mathcal{E}^i_j)_{j\in\N{}}$ and $(X^i_j)_{j\in\N{}}$ are
  independent.
\item $E\left(\mathcal{E}^{i}_{j}\right)=0$ and
  $E\left(\left|\mathcal{E}^{i}_{j}\right|^q\right)<\infty$, where $q$ is the
  conjugate exponent to $p$ used in hypothesis $H_c$ (2-c).
\end{enumerate}
For each $i$, the sequence $(X^i_j)_{j\in\N{}}$ corresponds to observation
points for the function $G^i$ and the sequence $(\mathcal{E}^i_j)_{j\in\N{}}$
corresponds to measurement errors for these observation points. More
precisely, if $g^i$, $x^i_j$ and $\varepsilon^i_j$ are respectively
realizations of $G^i$, $X^i_j$ and $\mathcal{E}^{i}_{j}$, we assume that we
observe the sequence: $y^i_j=g^i(x^i_j)+\varepsilon^i_j$. Moreover, we assume
that we know only the $m^i$ first values of this sequence. 

Hypothesis $H_e$ are natural in this framework, especially independence. The
main hypothesis is $H_e$ (2), which says that the way observation points are
randomly chosen (i.e., $P_X$) corresponds to the way integrals are calculated
($\mu$). On an intuitive point of view, this means that when an input function
is matched to functional weights thanks to integral calculation, probable
observation points have more weight that less probable ones. This is quite
natural.

As functions are only known thanks to observations, we cannot compute anymore
the integrals which are approximated thanks to empirical means. More
precisely, we replace $\int F_l\left( w_l,x\right) g^i(x)\td\mu(x)$ by:
\begin{equation}\label{eq:ApproxIntegral}
\frac{1}{m^i}\sum_{j=1}^{m^i}F_l(w_l,x^i_j)y^i_j.
\end{equation}
Therefore, the empirical error $\widehat{\lambda}_n(w)$ given in equation
\ref{eq:EmpiricalError} is approximated by the following empirical error:
\begin{multline}\label{eq:EmpiricalErrorDisc}
\lambda_{n}^{m}(w)=\\
\frac{1}{n}\sum_{i=1}^nc\left(t^i,U\left(w_o,\frac{1}{m^i}\sum_{j=1}^{m^i}F_1(w_1,x^i_j)y^i_j,\ldots,\frac{1}{m^i}\sum_{j=1}^{m^i}F_k(w_k,x^i_j)y^i_j\right)\right),
\end{multline}
where $t^i$ is a realization of $T^i$ and $m=\inf_{1\leq i\leq n} m^i$.

This empirical error, which is based on finite number of numerical values, is
easy to evaluate in practice and can be used to obtain empirical optimal
parameters, $w^m_n$. Our goal is to show that $w^m_n$ is a consistent
estimator of an optimal parameter vector, i.e. converges to $W^*$.

\subsection{Consistency}
Consistency of the proposed estimation of optimal parameters is given by the
following theorem:
\begin{thm}\label{TheoremFullConsistency}
Under hypothesis $H_b$, $H_c$, $H_d$ and $H_e$, we have $P$-almost surely:
\[
\lim _{n\rightarrow \infty }\lim _{m\rightarrow \infty
  }d(w_{n}^{m},W^{*})=0.
\]
\end{thm}
The theorem is an extension of \citet{White89Review}. It suffers from a small
limitation: the limit is a sequential one, which means that in order to reach
a given distance to $W^*$, the number of evaluation points for each function
($m$) depends on the number of functions ($n$). 

\section{Alternative methods for functional inputs}\label{sectionAlternative}
\subsection{Functions observed at identical points}
In some particular cases, functions are all observed thanks to an unique
sequence of observation points, that is there is a sequence
$(x_j)_{j\in\N{}}$ such that for any considered function $g$, we know $g(x_j)$
for all $j$. Moreover, we assume that we use the same number of observation
points for each function (denoted by $m$). These cases include for instance
situations in which measurement points are under user control (e.g.,
spectroscopic measurements corresponding to specific frequencies). Or course,
this case is covered by theorem \ref{TheoremFullConsistency}. 

On a practical point of view, the situation is clearly simpler than the
general one. Indeed each function $g$ can be considered as a vector in
$\R{m}$, i.e., $(g(x_1),\ldots,g(x_m))$. Therefore, we can submit these
multivariate observations to a numerical MLP. This approach was proposed in
\citet{ChenChen1995b}. Let us consider the special case of a single hidden
layer perceptron with one real output. Such a MLP maps a function $g$ to:
\begin{equation}
  \label{eqMLPChenChen}
V(g)=\sum_{i=1}^ka_iT\left(b_i+\sum_{j=1}^mc_{ij}g(x_j)\right).
\end{equation}
In such a setting, our model maps $g$ to:
\begin{equation}
  \label{eqFMLPUnique}
H(g)=\sum_{i=1}^ka_iT\left(b_i+\frac{1}{m}\sum_{j=1}^mF_i(w_i,x_j)g(x_j)\right).
\end{equation}
On a practical point of view, the main advantage of our approach over the
numerical one in this setting is the increased flexibility induced by the use
of the parametric functions $F_i$. We can for instance take into account
smoothness of observed functions by using simple parametric functions (i.e.,
MLP with a small number of hidden nodes, B-splines with just a few nodes,
etc.). This allows to reduce the number of free parameters in the model while
incorporating expert knowledge into it, whereas in the numerical
approach, we need in each neuron one connection weight for each function
observation point. 

Moreover, it is obvious that an appropriate choice of parametric functions
$F_i$ allows to reproduce exactly the numerical model, which appears this way
as a special case of the functional approach. Indeed each $F_i$ can be an
interpolation spline or a kernel based model designed such that for any set of
weights $c_{ij}$, there are weight vectors $w_i$ such that
$F_i(w_i,x_j)=c_{ij}$. 

Finally, the universal approximation result given in \citet{ChenChen1995b} is
less general than ours as it relies on uniform sampling. 

For all those reasons, we believe that the functional approach is more
interesting than the multivariate approach, even for uniformly sampled
functions. Experiments exposed in section \ref{sectionExperiments}
confirm this point of view.

\subsection{Function representation}\label{subsectionFunctionRepresentation}
When functions are not observed at identical evaluation points, there is still
a natural alternative approach to ours. The main idea is simply to transform
each functional observation into a representation that allows easy manipulation. 
More precisely, a list of observations
$(x_j,g(x_j)+\epsilon_j)_{j\in\{1,\ldots,m\}}$ is replaced by an approximation
of $g$, $A(g)$ constructed thanks to the observations. 

The only reasonable solution is to use a pseudo-linear model to approximate
the input/output mapping for each observed function. Indeed, the number of
input functions can be quite large in real world experiments and fitting a non
linear model to each function will be very time consuming. Morever, the only
difference between $A(g)$ and $g$ is that the former is known exactly whereas
the latter is not. Representation does not solve the function manipulation
problem. If we use non linear models, calculation of a scalar product between
$A(g)$ and a weight function is still a complex problem that cannot be solved
without an approximation method for integral calculation. We are more or less
back to our original problem, except that we have now perfectly known
functions (hopefully smoothed by the representation algorithm). We do not
discuss this approach anymore because it is in fact an extended version of our
method (whose theoretical properties remain to be studied).

The case of pseudo-linear models allows to construct what might be seen as an
alternative to our approach. Indeed, $A(g)$ is obtained thanks to a truncated
basis expansion, a very common approach in FDA thoroughly illustrated in
\citet{RamseySilverman97} and more recently in \citet{BesseCardotLivre2003}.
First of all, we need to assume that studied functions belong to $L^2(\mu)$.
We chose a free system of $L^2(\mu)$, $(\phi_i)_{1\leq i\leq p}$. Then each
list of observations $(x_j,g(x_j)+\epsilon_j)_{j\in\{1,\ldots,m\}}$ is
replaced by $A(g)$ the projection of $g$ on the vectorial space spanned by
$\phi_1,\ldots,\phi_p$, denoted $span(\Phi_p)$. On a practical point of view,
we simply calculate numerical parameters $\alpha_i(g)$ that minimize
\[
\sum_{j=1}^m\left(g(x_j)+\epsilon_j-\sum_{i=1}^p\alpha_i(g)\phi_i(x_j)\right)^2.
\]
This approach has two advantages over the general non linear representation
technique. First it is faster as $\alpha_i(g)$ is obtained very efficiently
thanks to some simple linear algebra. Second it can lead to a simplify neural
model. Indeed we can submit the numerical vector that represents a function
($(\alpha_i(g))_{1\leq i\leq p}$ to a numerical MLP (even if the observation
points depend on the function, because $p$ is the same for all functions). 

On a theoretical point of view, this solution is in fact a particular case of
our approach. Indeed our approach is based on calculating an approximation of
$\int fg\td\mu$. In $L^2(\mu)$, this is the scalar product. Let us consider the
special case where we constraint weight functions $f$ to belong to
$span(\Phi_p)$, i.e., $f=\sum_{i=1}^p\beta_i(f)\phi_i$. We have obviously
\[
\int fg\td\mu=\sum_{i=1}^p\beta_i(f)\int g\phi_i\td\mu.
\]
If we knew the real projection of $g$ on $span(\Phi_p)$, $\Pi(g)$, we would be
able to replace $\int g\phi_i\td\mu$ by $\int \Pi(g)\phi_i\td\mu$. This is not the
case, but we can still assume that $\int g\phi_i\td\mu$ is approximately equal
to $\sum_{j=1}^p\alpha_j(g)\int \phi_j\phi_i \td\mu$.  Therefore $\int fg\td\mu$
is approximately equal to $\sum_{i=1}^p\sum_{j=1}^p\beta_i(f)\alpha_j(g)\int
\phi_j\phi_i \td\mu$. Let us denote $M$ the matrix $M_{ij}=\int \phi_i\phi_j
\td\mu$. As $(\phi_i)_{1\leq i\leq p}$ is a free system, $M$ is a full rank
matrix. If we denote $\gamma(f)=M\beta(f)$, we have 
\[
\int fg\td\mu\simeq \sum_{j=1}^p\gamma_j(f)\alpha_j(g).
\]
Moreover, given a vector of coefficients $c$, we can define a function $t$ by
\[
t=\sum_{i=1}^pd_i\phi_i,
\]
with $d=M^{-1}c$ such that
\[
\int tg\td\mu\simeq \sum_{j=1}^p\gamma_j(t)\alpha_j(g)=\sum_{j=1}^p c_j\alpha_j(g).
\]
Therefore, a linear combination of the (approximate) coordinates of $g$ on
$span(\Phi_p)$, is always approximately equal to the scalar product of $g$
with a well chosen weight function $f$. Our method approximates $\int fg\td\mu$
by another formula. It is obvious that for the limit case, we end up with
identical values and therefore that our approach contains as a special case
the representation based approach. As in the previous section, this might be
even clearer with a simple one hidden layer perceptron with an unique real
output. The representation based approach maps $g$ to
\begin{equation}
V(g)=\sum_{i=1}^ka_iT\left(b_i+\sum_{l=1}^pc^i_{l}\alpha_l(g)\right),
\end{equation}
whereas our model gives 
\begin{equation}
H(g)=\sum_{i=1}^ka_iT\left(b_i+\frac{1}{m}\sum_{j=1}^m\left(g(x_j)+\epsilon_j\right)\left(\sum_{l=1}^pd^i_{l}\phi_l(x_j)\right)\right).
\end{equation}
According to the previous discussion, to obtain nearly identical values, we
just have to choose $d$ such that $d^i=M^{-1}c^i$, for all $i$ . Of course, on
a numerical point of view, results might be slightly different (as will be
illustrated in the following section), but the truncated basis approach can
still be considered as a different implementation of a special case of our
approach. More sophisticated truncated basis approaches, involving for
instance a roughness penalty as in \citet{BesseCardotFerraty1997}, depart more
from the solution proposed here and should be studied independently. 

\section{Experiments}\label{sectionExperiments}
\subsection{Introduction and experimental setting}
In the present section, we illustrate the proposed approach on two
supervised classification experiments. The first dataset, studied in section
\ref{subsectionBreiman}, consists in the traditional waveform data introduced
in \citet{Breiman84}. In this synthetic example, the goal is to classify
examples into three classes. The second dataset, studied in section
\ref{subsectionSpectro}, consists in a real world spectrometric problem in
which near infrared absorbance spectra are used to recognize high
fat and low fat meat samples. 

Both datasets have been used in \citet{FerratyVieu2002CSDA} to illustrate the
efficiency of the non-parametric functional kernel based model proposed in the
corresponding paper (and also in \citet{FerratyVieu2002CS}). We will therefore
compare results obtained thanks to neuronal approaches to functional and
classical methods used in \citet{FerratyVieu2002CSDA}. Those methods include
the above mentioned kernel based model as well as the linear model,
Partial Least Square Regression, CART, etc. 

We have considered three variations of the Multi Layer Perceptron: the
classical MLP applied to raw data, the functional approach presented in this
paper and the alternate implementation of the functional approach based on
projection on a B-spline basis (see section
\ref{subsectionFunctionRepresentation}).  

In all our experiments, we have used a conjugate gradient training algorithm,
with 10 different random initializations. To avoid over-fitting, we used a
weight decay penalization term. To select both the architecture of the MLP and
the value of the weight decay constant, we have used $k$-fold cross-validation
(with $k=5$). Finally, performances of the selected MLP have been evaluated on
a test sample. 

\subsection{Breiman waves}\label{subsectionBreiman}
\subsubsection{Classification results}
We start our experiments with synthetic data, more precisely with waveform
data introduced in \citet{Breiman84}. This is a three-class problem in which
each class is obtained thanks to convex combination of three shifted triangular
waveforms. The generating waveforms are continuous curves defined on $[1,21]$
by:
\begin{eqnarray}\label{eqWavesGenerator}
h_1(t) &=&\max(6-|t-11|,0),\\
h_2(t) &=&h_1(t-4),\\  
h_3(t) &=&h_1(t+4).
\end{eqnarray}
Functions to classify have the following general forms:
\begin{eqnarray}\label{eqWaves}
x(t) &=&uh_1(t)+(1-u)h_2(t)\ \ \text{for class 1,}\\
x(t) &=&uh_1(t)+(1-u)h_3(t)\ \ \text{for class 2,}\\
x(t) &=&uh_2(t)+(1-u)h_3(t)\ \ \text{for class 3,}
\end{eqnarray}
where $u\in ]0,1[$. In \citet{Breiman84} each function is transformed into a
vector from \R{21} thanks to an uniform sampling on $[1,21]$. An independent
standard Gaussian noise is added to each observation. 

In order to stay closer to the functional framework, we follow
\citet{FerratyVieu2002CSDA} and work therefore with vectors from \R{101} which
correspond to an uniform sampling of each function on $[1,21]$. The training
sample is obtained exactly as in \citet{FerratyVieu2002CSDA} : we have 150
functions in each class (in order to build such a function the parameter $u$
is chosen uniformly in $]0,1[$, independently for each function) and an
independent standard Gaussian noise is added to each observation. The test
sample is generated with the same method but contains 250 functions in each
class. 

As explained in the introduction, we have compared three neuronal approaches:
a naive approach in which \R{101} vectors are directly submitted to a
classical one hidden layer perceptron, our functional approach in which
functional weights are represented thanks to B-splines and the alternate
implementation of our method based on projection on the same B-splines basis.
We refer to \citet{FerratyVieu2002CSDA} for comparison with classical methods
and the non parametric functional method introduced in the paper. Table
\ref{tableBreimanResults} gives the obtained results for the three neural
methods (MLP corresponds to the naive approach, FMLP to our functional
approach and FpMLP to the alternate implementation of this approach).  Results
have been averaged over 50 simulations, exactly as in
\citet{FerratyVieu2002CSDA}, so as to ease comparison with existing results.
\begin{table}[htbp]
  \begin{center}
    \begin{tabular}{|l|c|c|}\hline
Method & Test classification error rate & Standard deviation \\\hline
MLP & 0.098 & 0.013\\\hline
FMLP & 0.065 & 0.0096\\\hline
FpMLP & 0.072 & 0.011\\\hline
    \end{tabular}
    \caption{Waveform data}\label{tableBreimanResults}
  \end{center}
\end{table}

Results are very satisfactory. First of all, our functional approaches
overcome the classical MLP method (the main functional approach gives the best
results). Result summary provided by table \ref{tableBreimanResults} does not
give complete information. Indeed, as for each simulation the same data set is
used for each method, a direct comparison between obtained results is
possible. An important result is that for \emph{all} simulations, functional
approaches overcome the classical MLP. The mean performance increase is 3.2
percent for our main implementation and 2.6 percent for the alternate
projection based implementation. Moreover, the main implementation overcomes
the projection based one on 38 simulations (and the mean performance increase
is 0.6 percent).

According to results reported in \citet{FerratyVieu2002CSDA}, the functional
MLP approach outperforms both classical methods (such as CART) and functional
ones. The best method studied in \citet{FerratyVieu2002CSDA} achieves a mean
classification error rate of $0.072$ (with a standard deviation of
$0.012$). We can therefore conclude that our functional MLP is among the best
methods for this dataset and that it overcomes both traditional methods and a
classical neural approach. Moreover, as explained in the following section,
the obtained functional model is very parsimonious which gives it robustness
and efficiency. 

\subsubsection{Parameter numbers}
For all methods, we select the best number of hidden neurons among
2, 3 and 4 hidden neurons. For the functional approaches, weight
functions were represented using 5, 7, 10, 15 or 20 B-splines (those numbers
have been chosen to keep the architectures as simple as possible). The chosen
architecture depends on the simulation, but in general, small architectures
are preferred, as summarized by the following tables. Table \ref{tableBspline}
gives the number of time each B-splines basis has been chosen and table
\ref{tableNeurons} gives the number of time each number of hidden neurons has
been chosen.
\begin{table}[htbp]
  \centering
  \begin{tabular}{|l|c|c|c|c|c|}\hline
Number of B-splines & 5 & 7 & 10 & 15 & 20 \\\hline
FMLP & 18 & 17 & 8 & 3 & 4 \\\hline
FpMLP & 38 & 8 & 4 & 0 & 0 \\\hline
  \end{tabular}
  \caption{Number of simulations that select the given number of B-splines}%
\label{tableBspline}
\end{table}

\begin{table}[htbp]
  \centering
  \begin{tabular}{|l|c|c|c|}\hline
Number of hidden neurons & 2 & 3 & 4 \\\hline
MLP & 10& 40 & 0 \\\hline
FMLP & 9 & 29 & 12 \\\hline
FpMLP & 10 & 28 & 12  \\\hline
  \end{tabular}
  \caption{Number of simulations that select the given number of hidden neurons}%
\label{tableNeurons}
\end{table}

For our main functional approach, the total number of numerical parameters used
varies between 23 and 103, with a mean of 44 (the median is 39 and only 10
simulations needed more than 51 parameters). For the projection based
implementation, the total number of numerical parameters used varies between 23
and 63, with a mean of 36 (the median is 33 and only one simulation out of 50
uses more than 51 parameters). The projection based approach uses therefore
even less parameters than our main functional approach, but with a slight
decrease in the performances.

For the naive approach, cross-validation selects 3 hidden neurons for 10
simulations and 4 hidden neurons for the other 40 simulations. Those values
correspond respectively to 321 and 427 numerical parameters (the mean is
406). The naive approach uses therefore far more parameters than functional
methods and gives worse results. 

The best method studied in \citet{FerratyVieu2002CSDA} is a non-parametric
functional method in which functions are first projected on an optimal basis
constructed thanks to multivariate partial least squares regression. Optimal
results are obtained thanks to a projection on three basis functions (this
number is selected thanks to $k$-fold cross-validation). As the method is
kernel based, we have to store all the functions of the training sample. That
is, we need to keep a vector of \R{101} for each basis function (303 numerical
parameters) as well as the coordinate of each training function of this basis
(3 parameters for each function). We have therefore a total of 1653 numerical
parameters.

\subsubsection{Pre-smoothing}\label{subsubsectionPreSmooth}
A possible explanation for the poor performances of the standard MLP is that
Breiman waves are very noisy. One side effect of using function
representation, either for the functional weights or for the data themselves,
is to smooth the waves. It is therefore quite natural to investigate the
effect of applying a spline smoothing method on the waves before submitting
them to a standard MLP. 

In order to implement a fair comparison, we have used the following method: we
calculate coordinates of training and test waves on each B-spline basis
considered in the previous series of experiments. These coordinates are used
to reconstruct smooth versions of the waves that are sampled exactly as the
original waves (101 points regularly spaced in $[1,21]$). The obtained \R{101}
vectors are then submitted to a classical one hidden layer perceptron. The
number of B-splines used for the smoothing phase, the number of hidden neurons
and the weight decay are then selected by $k$-fold cross validation (with
$k=5$).

The test set performances are much better than with the basic MLP approach.
Indeed, the mean error rate is now $0.073$ (with a standard deviation of
$0.012$), which is comparable to the non-parametric approach of
\citet{FerratyVieu2002CSDA} and to the projection based implementation of the
functional MLP. Nevertheless, a direct comparison shows that in fact our main
implementation performs better than the smoothing approach for 42 simulations
on 50. The projection based implementation obtains better results for 27
simulations. The basic MLP approach obtains better results than the smoothing
approach for 1 simulation out of 50.

It seems therefore that smoothing plays an important role in obtaining good
performances, but also that it does not help in reducing the number of
parameters. Indeed, the mean number of parameters used by the smoothing
approach is 406. With only 44 parameters, our main implementation obtains
slightly better results.

\subsubsection{Comments}
Table \ref{tableBreimanSummary} summarizes the result obtained on the Breiman
waves. It is clear that the functional MLP approach gives very satisfactory
results on those data. The obtained classification rate is slightly better
than the best results reported in \citet{FerratyVieu2002CSDA}, which means
that the MLP approach performs better than both traditional approaches and
functional approaches. Moreover, the functional MLP approach also overcomes a
naive MLP modeling of the raw multivariate data, as well as a more complex
method in which a spline smoothing is performed on the raw data before
submitting them to a classical MLP. Finally, the obtained model is very
parsimonious: the MLP classifier will be faster than the kernel based one
(after training).

\begin{table}[htbp]
  \centering
  \begin{tabular}{|l|c|c|c|}\hline
Model & Parameters & Error rate & Training time\\\hline
FMLP & 44 & 0.065&4.5\\\hline
FpMLP & 36& 0.072&3.8\\\hline
Non parametric & 1653 & 0.072&0\\\hline
MLP with smoothing & 406& 0.073&5.9\\\hline
MLP & 406 & 0.098&1\\\hline
  \end{tabular}
  \caption{Results summary}%
\label{tableBreimanSummary}
\end{table}

Table \ref{tableBreimanSummary} shows also the relative cost of the studied
methods in terms of training time: the total training time of the classical
MLP applied on raw data has been chosen as the reference training time (the
values include the cross validation phase).  As the non parametric approach
of \citet{FerratyVieu2002CSDA} involves almost no training phase (except for
the selection of the kernel width), it has been considered as almost
instantaneous compared to MLP training. Most of the cost comes from the model
selection phase. Indeed, for the basic MLP, we just have to select the weight
decay parameter and the number of hidden neurons. On the contrary, all other
methods involve the selection of the representation basis (here the number of
B-splines). An interesting point is that the functional approaches are faster
to train than the smoothing approach, give better results and produce very
parsimonious summary of the data.

Compared to a classical MLP, the functional approach implies to use around 4.5
times more processing power in the training phase.  Fortunately, the training
is done only once and allows to produce a very small footprint solution than
can be implemented on a small device such as a cell phone or a PDA, and with
recognition performances that are significantly better than those of the
classical MLP.

\subsection{Spectrometric data}\label{subsectionSpectro}
\subsubsection{Raw data}
Our next example is a real world classification problem of spectrometric data
from food industry. Each observation is the near infrared absorbance spectrum
of a meat sample (finely chopped), recorded on a Tecator Infratec Food and
Feed Analyser. More precisely, an observation consists in a 100 channel
spectrum of absorbances in the wavelength range 850--1050 nm. The goal is to
classify meat samples into high fat samples and low fat samples. The first
class consists in meat samples with less than 20\% of fat, whereas the second
class contains all other meat samples. We have a total of 215 spectra. Data
are not organized into a training sample and a test sample, therefore, we
follow exactly the evaluation method described in \citet{FerratyVieu2002CSDA}:
we select randomly 160 training spectra and 55 test spectra. We repeat
this operation 50 times and give the average classification error rate.

We have compared the three approaches described in the introduction. The
preprocessing experimented in section \ref{subsubsectionPreSmooth} was not
considered here because absorbance spectra are very smooth and a B-spline
basis projection as no noticeable smoothing effect on those functions. Table
\ref{tableSpectroResults} gives statistical summaries of the classification
error rate obtained by those neural methods.
\begin{table}[htbp]
  \begin{center}
    \begin{tabular}{|l|c|c|c|c|}\hline
Method & First quartile & Mean & Median & Third quartile\\\hline
MLP & 0 & 0.019 & 0.018&0.036\\\hline
FMLP & 0.018 &0.028 & 0.036&0.036\\\hline
FpMLP &0 & 0.018 & 0.018&0.036\\\hline
    \end{tabular}
    \caption{Error rate for Spectrometric curves}\label{tableSpectroResults}
  \end{center}
\end{table}
In this situation, only the alternate implementation of the functional
approach gives satisfactory results. Indeed, the naive MLP approach gives
better results than our main functional implementation. As in the previous 
section, generated data sets are identical for each method and a direct
comparison between obtained results is possible. The naive method performs
better than the FMLP method on 21 data sets (identical performances are
obtained on 27 simulations). 

But the FpMLP method still performs better than the naive approach. The
average performance improvement is only 0.001, but FpMLP performs better than
MLP on 30 simulations (identical performances are obtained on 13
simulations). We can therefore conclude that the best functional approach
gives slightly better performances than the MLP approach. Moreover,
the best method reported in \citet{FerratyVieu2002CSDA} obtains
a median classification rate of approximately $0.022$, which shows again that
neural methods perform very well. Additionally, the best method reported in
\citet{FerratyVieu2002CSDA} is as in previous section a mixed method that uses
a functional non parametric model on functions projected on an optimal basis
generated thanks to non functional multivariate partial least squares
regression. \citet{FerratyVieu2002CSDA} reports that a pure functional
approach (in which functional principal component analysis is used to design
an optimal projection) gives very bad results (the mean error rate is
$0.2$). On the contrary, our methods are pure functional methods and still give
the best results. 

Moreover, functional methods use a small number of numerical parameters. For
all methods, we select the best number of hidden neurons among 2, 3 and 4
hidden neurons. For the functional approaches, weight functions were
represented using 15 or 20 B-splines. In general, methods choose a small
number of neurons, as shown in tables \ref{tableBsplineSpectro}
and \ref{tableNeuronsSpectro}.

\begin{table}[htbp]
  \centering
  \begin{tabular}{|l|c|c|c|c|c|}\hline
Number of B-splines & 15 & 20 \\\hline
FMLP & 12 & 38   \\\hline
FpMLP & 24 & 26  \\\hline
  \end{tabular}
  \caption{Number of simulations that select the given number of B-splines}%
\label{tableBsplineSpectro}
\end{table}

\begin{table}[htbp]
  \centering
  \begin{tabular}{|l|c|c|c|}\hline
Number of hidden neurons & 2 & 3 & 4 \\\hline
MLP & 24& 16 & 10 \\\hline
FMLP & 32 & 11 & 7 \\\hline
FpMLP & 18 & 17 & 15  \\\hline
  \end{tabular}
  \caption{Number of simulations that select the given number of hidden neurons}%
\label{tableNeuronsSpectro}
\end{table}

The classical MLP approach uses between 213 and 423 parameters, with a mean of
288 parameters. The main functional approach uses between 43 and 103
parameters (the mean is 62), whereas the projection based approach has the
same range of parameter numbers with a higher mean (69). The best method
reported in \citet{FerratyVieu2002CSDA} uses 1300 parameters (almost 19 times
more than our best method) with slightly worse performances.

\subsubsection{Second order derivatives}

\citet{FerratyVieu2002CS} and \citet{FerratyVieu2002CSDA} point out that
the second derivative of the spectrum is in general more informative than the
spectrum itself. The non parametric approach proposed in
\citet{FerratyVieu2002CSDA} has been used with a second derivative based
semi-metric and achieved better results than the optimal projection based
method. Indeed, the median error rate of a pure functional approach is now
slightly less than $0.022$. This method turns out to be the best overall
method. 

We have therefore applied our functional MLP approaches to the second
derivative of the spectrum. As in \citet{FerratyVieu2002CSDA}, we evaluate the
spectrum thanks to a B-spline representation. The second derivative of the
B-spline is calculated exactly and sampled uniformly on $[850,1050]$ as the
original data. We obtain therefore new functional data that we model as normal
functional data (that is we forget the preprocessing phase).

Table \ref{tableSpectroDerivResults} gives statistical summaries of the
classification 
error rate obtained by the neural methods applied to the second order
derivatives. 

\begin{table}[htbp]
  \begin{center}
    \begin{tabular}{|l|c|c|c|c|}\hline
Method & First quartile & Mean & Median & Third quartile\\\hline
MLP & 0 & 0.013& 0.018 & 0.018\\\hline
FMLP & 0 & 0.007& 0 & 0.018\\\hline
FpMLP &0 & 0.014&0.009 & 0.018\\\hline
    \end{tabular}
    \caption{Error rate for second order derivatives of the Spectrometric
      curves}\label{tableSpectroDerivResults} 
  \end{center}
\end{table}

We obtain very satisfactory results as all neural methods perform better than
results reported in \citet{FerratyVieu2002CSDA}. Moreover, the best results
are obtained by our main functional MLP implementation. A direct comparison
between results obtained for each simulation shows that FMLP overcomes MLP on
15 simulations (identical performances are obtained on 31 simulations). The
FMLP also overcomes FpMLP on 19 simulations (identical performances are
obtained on 20 simulations). In fact FMLP provides perfect classification of
the test set for 34 simulations, whereas this number drops to 25 for FpMLP and
to 24 for MLP. 

We do not report completely architecture selection results as they are very
similar to those obtained on the raw functional data. MLP uses a mean number
of 391 parameters, FMLP 82 and FpMLP 76. 

\subsubsection{Comments}
As in the Breiman wave experiments, an appropriate functional MLP model allows
to obtain very good recognition rate that cannot been reached by a classical
MLP. Moreover, the optimal functional MLP uses a small number of parameters,
which eases its real world implementation. We have not reported here training
times as they are comparable to values reported in table
\ref{tableBreimanSummary}: the price to pay for higher recognition rate and
lower parameter number is a higher training time than the one needed for a
classical MLP, mainly because of the additional parameter (the number of
B-splines) that has to be chosen by cross validation. 

\subsection{Conclusions}
In both experiments (on simulated data and on real world data), functional
multi-layer perceptrons perform in a very satisfactory way. They are at least
as good as functional and traditional methods presented in
\citet{FerratyVieu2002CSDA}. Moreover, they also overcome a naive MLP modeling
of the raw multivariate data. A way to obtain correct results with a classical
MLP is to perform a kind of functional preprocessing: a spline smoothing for
noisy data such as the Breiman wave or a derivative calculation for smooth
data such as the absorbance spectra. But even those mixed approaches do not
perform as well as the functional MLP. Another important practical property is
the small number of numerical parameters used by the functional neural
methods: this allows an easier implementation on devices with limited
resources such as PDA, cell phones and more generally embedded devices. 

Of course, additional experiments on real world data are needed to fully
understand advantages and shortcomings of the proposed functional MLP. While
the model has been compared to traditional classification methods thanks to
experiments conducted in \citet{FerratyVieu2002CSDA}, additional comparisons,
especially to recent methods such as support vector machines (see e.g.
\citet{ChristianiniShaweTaylor2000SVMIntroduction}) or boosted classification
trees (see e.g. \citet{HastieTibshiraniFriedman2001SL}), are also needed.

An interesting open research topic is to develop automatic tuning of weight
function representation. We have used here a brute force $k$-fold
cross-validation method but \citet{FerratyVieu2002CSDA} shows that automatic
design of projection basis can improve performances. Moreover, this might
reduce the training time of functional MLP which remains the only negative
part of the proposed approach compared to classical MLP (when the latter is
used without functional preprocessing).

\section{Conclusion}
In this paper, we have introduced Functional Multi-Layer Perceptrons (FMLP), a
simple extension of MLP to functional data. The proposed model is very
interesting on a theoretical point of view because it shares with its numerical
counterpart useful properties. 

We have indeed shown that FMLP are universal approximators, that is they can
approximate continuous mappings from a compact subset of a functional space to
\R{} with arbitrary precision. For a given function to approximate to a given
accuracy, the approximating FMLP uses a finite number of numerical parameters.

Moreover, we have shown that parameter estimation for FMLP is consistent:
optimal parameters estimated thanks to a finite number of functions known at a
finite number of measurement points converge to the set of true optimal
parameters when the size of the data increases. 

We have also shown on simulated and real world data that the FMLP performs in
a very satisfactory way. Performances are in general better than those
obtained by non functional methods (including neural methods) and at least as
good as other functional methods. Moreover, the functional approach gives much
more parsimonious representation of studied data, a property that enhance the
robustness of the obtained models and allows also an easier implementation on
devices with limited processing power. We believe therefore that Functional
Multi-Layer Perceptrons are a valuable tool for data analysis when a
functional representation of input variables is possible.

\section*{Acknowledgement}
The authors thank participants to the working group STAPH
(\url{http://www.lsp.ups-tlse.fr/Fp/Ferraty/staph.html}) on Functional
Statistics at the University Paul Sabatier of Toulouse, for very interesting
and stimulating discussions. The authors thank especially Fr\'ed\'eric Ferraty
and Philippe Vieu for sharing data and simulation results. The authors 
thank also anonymous referees for their valuable suggestions that help
improving this paper.

\section{Proofs}\label{sectionProofs}
\begin{pf*}{Proof of corollary \ref{corollary:symbolicNN:universal}}
If $ 1<p<\infty  $, we know that $ L^{q}(\mu ) $ (with $ q<\infty ) $
can be identified with $ \left( L^{p}(\mu )\right) ^{*} $ (see
for instance \citet{RudinRealAndComplexAnalysis}). More precisely,
for each $ l\in \left( L^{p}(\mu )\right) ^{*} $ there is an unique
function $ f\in L^{q}(\mu ) $ so that $ l(g)=\int fg\, \td\mu  $. By hypothesis,
$V$ is dense in $L^{q}(\mu)$. This obviously implies that $A_V$ is dense in $
\left( L^{p}(\mu )\right) ^{*}$ for 
the weak $ * $ topology. We can therefore apply corollary 5.1.3 of
\citet{Stinchcombe99} (note that corollary 5.1.3 is given for the outside
density case, but the author states explicitly that a similar inside
corollary is valid). 

If $ p=\infty $, we cannot apply directly corollary 5.1.3 from
\citet{Stinchcombe99} as the dual of $ L^{\infty }(\mu ) $ is not $ L^{1}(\mu
) $.  Let us nevertheless consider $ A $ the set of afinne functions on $
L^{\infty }(\mu ) $ defined by $ l(f)=\alpha +\int fg\, \td\mu $, where $ \alpha
$ is an arbitrary real number and $ g $ is an arbitrary function from
$V\subset L^1(\mu)$. $ A $ is obviously a vectorial space which contains
constant functions of $ C(K,\R{} ) $.  Let us now show that $ A $ separates
points in $ K $. Let $ u $ and $ v $ be two distinct functions of $ K $. The
function $ f=u-v $ is a non zero function belonging to $ L^{\infty }(\mu ) $.
We can assume that the measurable set $ H=\{x\in \R{n}\mid f(x)>0\} $ has non
zero finite measure (if it is not the case, replace $ f $ by $ -f $ ). Then,
obviously $ \int f\chi _{H}\td\mu >0 $, that is $ \int u\chi _{H}\td\mu \neq \int
v\chi _{H}\td\mu $. As $ \mu $ is finite, $ \chi _{H} $ belongs to $ L^{1}(\mu )
$. As $V$ is dense in $ L^{1}(\mu ) $, there is a sequence $ h_{k} $ of
functions in $V$ that converges to $ \chi _{H} $. We have obviously 
\[ 
\left|
  \int f(h_{k}-\chi _{H})\td\mu \right| \leq \left| f\right| _{\infty }\left|
  \int h_{k}-\chi _{H}\td\mu \right|.
\]

Therefore, there is an index $ k $ such that $ \int fh_{k}\td\mu >0 $, that is
there is a function $ h_{k}\in V $ such that $ \int uh_{k}\td\mu \neq \int
vh_{k}\td\mu $.  Therefore, $ A $ separates points in $ K $. The conclusion is
then obtained by applying theorem 5.1 of \citet{Stinchcombe99}.
\end{pf*}

\begin{pf*}{Proof of corollary \ref{corollary:symbolicNN:lone}}
  As $\mu$ is a finite Borel measure on $\R{n}$, it is regular
  (\citet{RudinRealAndComplexAnalysis}, theorem 2.18), and we can apply
  Lusin theorem (\citet{RudinRealAndComplexAnalysis}, theorem and corollary
  2.23). We know therefore that for any function $ f $ in $ L^{\infty }(\mu )
  $, there is a sequence of compactly supported continuous functions $ g_{k} $
  that converges punctually to $ f $ and such that $ |g_{k}|_{\infty }\leq
  |f|_{\infty } $.  A simple application of Lebesgue dominated convergence
  theorem shows that for any function $ h $ in $ L^{1}(\mu ) $, $ \int
  g_{k}h\, \td\mu \rightarrow _{k\rightarrow \infty }\int fh\, \td\mu $.  Then, as
  $ \mu $ is compactly supported, there is a compact $ K $ such that $ \int
  g_{k}h\, \td\mu =\int _{K}g_{k}h\, \td\mu $. Then, thanks to hypothesis, each $
  g_{k} $ can be approximated by a function $ \phi _{k} $ in $V$ such that $
  \sup _{x\in K}\left| g_{k}(x)-\phi _{k}(x)\right| <\frac{1}{k} $.  In this
  case $ \left| \int _{K}g_{k}h\, \td\mu -\int _{K}\phi _{k}h\, \td\mu \right|
  <\frac{1}{k}\left\Vert h\right\Vert _{1} $.  As $ \mu $ is compactly
  supported, this allows to conclude that $ \int \phi _{k}h\, \td\mu
  \xrightarrow[k\rightarrow \infty ]{}\int fh\, \td\mu $.  Therefore, the set of
  linear forms $A_V$ is dense for the weak $ * $ topology in $ \left(
    L^{1}(\mu )\right) ^{*} $, provided that $ \mu $ is finite and compactly
  supported. The conclusion is then obtained by applying corollary 5.1.3 from
  \citet{Stinchcombe99}.
\end{pf*}

\begin{pf*}{Proof of theorem \ref{TheoremFullConsistency}}
The proof is quite technical and can be cut into several parts:
\begin{enumerate}
\item We need first a quite general Uniform Strong Law of Large Numbers
  (USLLN) which will be obtained thanks to a general result of
  \citet{Andrews87}. 
\item Then we show that integral approximations used in the definition of
  $\lambda_{n}^{m}(w)$ have a kind of uniform convergence property.
\item Using both results, we show that $\lambda_{n}^{m}(w)$ converges
  almost surely uniformly to $\lambda(w)$.
\item The conclusion is obtained thanks to a simple lemma on approximation of
  the minimizers of a function.
\end{enumerate}
\textbf{part 1} 

A very general Uniform Strong Law of Large Numbers (USLLN) is given in
\citet{Andrews87}. It is based on complex assumptions, so we propose to
simplify it into the following corollary:
\begin{cor}\label{theoremUniformSLLN}
Let $ X $ be an arbitrary metric space considered with its Borel
sigma algebra. Let $ (\Omega ,\mathcal{A},P) $ be a probability
space on which is defined a sequence of independent identically distributed
random elements, $ Z_{t} $ with values in $X$. Let $ W $ be a compact metric space.
Let $ l $ be a function from $ W\times X $ to $ \R{}  $. We
assume that the following conditions hold:
\begin{enumerate}
\item For each $ w\in W $, $ l(w,.) $ is a measurable function from
$ X $ to $ \R{}  $.
\item For each $ x\in X $, $ l(.,x) $ is a continuous function from
$ W $ to $ \R{}  $.
\item there is a positive measurable function $ d $ (from $ X $ to
$ \R{}  $) such that for all $ x\in X $ and for all $ w\in W $,
$ |l(w,x)|\leq d(x) $. 
\item $ E(d(Z_{t}))<\infty  $.
\end{enumerate}
Then we have:\[
\sup _{w\in W}\left| \frac{1}{n}\sum ^{n}_{i=1}l(w,Z_{i})-E(l(w,Z_{t}))\right| \rightarrow ^{a.s.}_{n\rightarrow \infty }0.\]
\end{cor}
In order to prove this corollary, we need first a simple lemma:
\begin{lem}\label{lemma:sup:measurable}
Let $ l $ be a function from $ W\times X $
to $ \R{}  $, where $ W $ is a separable metric space and $ X $ is a metric space
(considered with its Borel sigma algebra). If
$ l $ is continuous on $ W $ for each fixed $ x\in X $ and
measurable on $ X $ for each fixed $ w\in W $, then the function
$ f(x)=\sup _{w\in W}l(w,x) $ is measurable.
\end{lem}
\begin{pf*}{Proof of lemma \ref{lemma:sup:measurable}}
As $ W $ is separable, there is a denombrable set $ W'=\{w_{i}\mid i\in \N{*}\}
$ 
dense in $ W $. Let us show that $ f(x)=\sup _{w\in W'}l(w,x) $.
Let us consider a fixed $ x\in X $. Let $ \epsilon  $ be an
arbitrary positive real number. By definition of $ f $, there is
$ w\in W $ such that $ l(w,x)\geq f(x)-\frac{\epsilon }{2} $.
As $ l(.,x) $ is continuous in $ w $, there is $ \eta  $
such that $ |w'-w|<\eta  $ implies $ |l(w',x)-l(w,x)|<\frac{\epsilon }{2} $,
which implies $ l(w',x)\geq f(x)-\epsilon  $. As $ W' $ is dense
in $ W $, there is $ w'\in W' $ such that $|w'-w|<\eta  $. This implies
$ f(x)\geq \sup _{w\in W'}l(w,x)\geq f(x)-\epsilon  $. As this
is true for each $ \epsilon  $, we have obviously $ f(x)=\sup _{w\in W'}l(w,x)
$. 
Therefore, $ f(x)=\sup _{i\in N}l(w_{i},x) $. As each function
$ l(w_{i},x) $ is measurable, the $ \sup  $ is also measurable.
\end{pf*}
We can now proceed to the proof of the corollary:
\begin{pf*}{Proof of corollary \ref{theoremUniformSLLN}}
We obtain corollary \ref{theoremUniformSLLN} as a consequence of Andrews'
theorem (\citet{Andrews87}). We have to check three assumptions:
\begin{enumerate}
\item Assumption $A\, 1$ is fulfilled as $W$ is compact ($W$ corresponds to
  $\Theta$ in Andrews' paper) 
\item Assumption $A\, 2$ breaks into two sub-assumptions:
  \begin{enumerate}
  \item Assumption $A\, 2\ (a)$ can be translated with our notation into the
    following assumption: for all $w_{0}$ (and all $i$), $l(w_{0},Z_i)$, $
    \sup _{w\in W(w_{0},\eta )}l(w,Z_i) $ and $ \inf _{w\in W(w_{0},\eta
      )}l(w,Z_i)$ are random variables (where $W(w_{0},\eta )=B(w_{0},\epsilon
    )\cap W$, and $B(w_{0},\epsilon)$ is the closed ball centered on $w_0$ with
    radius $\epsilon$).\newline
    
    $l(w_{0},Z_i)$ is a random variable thanks to assumption 1 of corollary
    \ref{theoremUniformSLLN}. Thanks to assumptions 1 and 2 of corollary
    \ref{theoremUniformSLLN} and due to the fact that a compact set is
    separable, lemma \ref{lemma:sup:measurable} can be applied to $l$ and to
    $W(w_{0},\eta)$, and allows to conclude that $\sup _{w\in
      W(w_{0},\eta )}l(w,Z_i) $ is a random variable. The case of $ \inf
    _{w\in W(w_{0},\eta)}l(w,Z_i)$ is handled thanks to the same
    lemma applied to $-l$.\newline

    Assumption $A\, 2\ (a)$ is therefore fulfilled. 
  \item Assumption $A\, 2\ (b)$ translates in our case into the assumption
    that $\sup _{w\in W(w_{0},\eta )}l(w,Z_i)$ and $\inf _{w\in W(w_{0},\eta
      )}l(w,Z_i)$ satisfy a point-wise strong law of large numbers, that is
    for any fixed $w_{0}$:
\[
\lim_{n\rightarrow\infty}\frac{1}{n}\sum_{i=1}^n\sup _{w\in W(w_{0},\eta
      )}l(w,Z_i)=E\left(\sup _{w\in W(w_{0},\eta
      )}l(w,Z_i)\right)\text{ $P$ a.s.}
\] \newline 

   As shown in the previous point, both $\left(\sup _{w\in W(w_{0},\eta
   )}l(w,Z_i)\right)_{i\in\N{*}}$ and $\left(\inf _{w\in W(w_{0},\eta
   )}l(w,Z_i)\right)_{i\in\N{*}}$ are sequences of independent identically
   distributed random variables. Moreover, thanks to assumptions 3 and 4 of
   corollary \ref{theoremUniformSLLN}, they are integrable and therefore the
   strong law of large numbers applies: assumption $A\, 2\ (b)$ is therefore
   fulfilled. 
  \end{enumerate}
\item Assumption $A\, 3$ translates in our case into the following assumption:
\[
\lim_{\eta\rightarrow 0}\sup_{n\geq 1}\left|\frac{1}{n}\sum_{i=1}^n
\left(E\left(\sup _{w\in W(w_{0},\eta)}l(w,Z_i)\right)-E\left(l(w,Z_i)\right)\right)\right|=0.
\]
A similar equation has to be fulfilled by $E\left(\inf _{w\in W(w_{0},\eta)}
l(w,Z_i)\right)$.\newline

As $l$ is continuous with respect to $w$ for a fixed $x$, we have the
following point-wise convergence:
\[
\lim_{\eta\rightarrow 0}\sup _{w\in W(w_{0},\eta)}l(w,.)=l(w_0,.).
\]
Thanks to assumptions 3 and 4 of corollary \ref{theoremUniformSLLN}, we can
apply Lebesgue dominated convergence which implies:
\[
\lim_{\eta\rightarrow 0}E\left(\sup _{w\in W(w_{0},\eta)}l(w,Z_i)\right)=
E\left(l(w_0,Z_i)\right).
\]
Finally, as $Z_i$ are identically distributed, assumption $A\, 3$ can be
simplified into:
\[
\lim_{\eta\rightarrow 0}\left|E\left(\sup _{w\in W(w_{0},\eta)}l(w,Z_1)\right)-E\left(l(w,Z_1)\right)\right|=0,
\]
which is exactly what we have just proven. The case of $E\left(\inf _{w\in
W(w_{0},\eta)} l(w,Z_i)\right)$ can be obtained exactly the same way.\newline

Assumption $A\, 3$ is therefore fulfilled.
\end{enumerate}
As the assumptions are fulfilled, we can apply Andrews' theorem which gives
exactly the conclusion of corollary \ref{theoremUniformSLLN}.
\end{pf*}

\textbf{part 2} 

Let us define:
\[
M^i_l(g,w_l)(\omega)_{m}=\frac{1}{m}\sum_{j=1}^mF_l(w_l,X^i_j(\omega))\left(g(X^i_j(\omega))+\mathcal{E}^i_j(\omega)\right),
\]
which can be simplified into $M^i_l(g,w_l)_m$ when $\omega$ is obvious, and
\[
M_l(g,w_l)=\int F_l(w_l,x)g(x)\td\mu(x)
\]
We prove now the following lemma:
\begin{lem}\label{lemma:unif:integ}
Let us define
\[
\Omega_l=\left\{\omega\in\Omega\mid\lim_{m\rightarrow\infty}\frac{1}{m}\sum_{j=1}^md_l(X^i_j)=\int
d_l\td\mu\right\}
\] and
\[
B^i_l=\left\{\omega\in\Omega\mid \forall g\in C(Z,\R{}),\
  \lim_{m\rightarrow\infty}\sup_{w_l\in
  W_l}\left|M^i_l(g,w_l)(\omega)_{m}-M_l(g,w_l)\right|=0 \right\}.
\]
Under hypothesis $H_a$, $H_c$ and $H_e$, $B^i_l\bigcap\Omega_l$ is measurable
and $P(B^i_l\cap\Omega_l)=1$. 
\end{lem}
\begin{pf*}{Proof of lemma \ref{lemma:unif:integ}}
The proof is based on the separability of $C(Z,\R{})$ and on corollary
\ref{theoremUniformSLLN}. Let us first note that $P(\Omega_l)=1$ thanks to
hypothesis $H_c$ (2-c) and the strong law of large numbers. 
Let us first show that the set
\[
B^i_l(g)=\left\{\omega\in\Omega\mid 
  \lim_{m\rightarrow\infty}\sup_{w_l\in
  W_l}\left|M^i_l(g,w_l)(\omega)_{m}-M_l(g,w_l)\right|=0 \right\}
\]
is such that $P(B^i_l(g))=1$ for any $g\in C(Z,\R{})$.

This can be obtained by applying corollary \ref{theoremUniformSLLN} to the
function $\psi$ from $W_l\times(Z\times\R{})$ to \R{} defined as follows
\[
\psi(w_l,(x,e))=F_l(w_l,x)(g(x)+e),
\]
and to the sequence of random elements $(X^i_j,\mathcal{E}^i_j)_{j\in\N{}}$. 
Corollary \ref{theoremUniformSLLN} applies because:
\begin{itemize}
\item $W_l$ is compact (hypothesis $H_c$ (1))
\item hypothesis $H_c$ (2-b) implies that $\psi$ is measurable with respect to
  its second variable
\item hypothesis $H_c$ (2-a) and $g\in C(Z,\R{})$ implies that $\psi$ is
  continuous with respect to its first variable 
\item hypothesis $H_c$ (2-c) implies
  $\forall x\in Z,\ w_l\in W_l$ and $\forall e\in\R{}$, $|\psi(w_l,(x,e))|\leq
  d_l(x)(|g(x)|+|e|)$ 
\item as $g$ is continuous on the compact set $Z$, $g\in L^q(\mu)$ and
  therefore $d_l(x)|g(x)|\in L^1(\mu)$ (according to hypothesis $H_c$ (2-c))
\item as $E\left(\left|\mathcal{E}^i_j\right|^q\right)<\infty$ (hypothesis
  $H_e$ (5)), $E(d_l(X^i_j)\left|\mathcal{E}^i_j\right|)<\infty$
\item $(X^i_j,\mathcal{E}^i_j)_{j\in\N{}}$ is i.i.d. (hypothesis $H_e$)
\end{itemize}
Therefore, we have:
\[
\sup_{w_l\in W_l}\left|M^i_l(g,w_l)_m-E\left(F_l(w_l,X^i_1)\left(g(X^i_1)+\mathcal{E}^i_1\right)\right)\right|\rightarrow ^{a.s.}_{m\rightarrow \infty }0.
\]
By definition, $E\left(F_l(w_l,X^i_1)g(X^i_1)\right)=M_l(g,w_l)$ and by
independence and hypothesis $H_e$ (5):
\[
E\left(F_l(w_l,X^i_1)\mathcal{E}^i_1\right)=E\left(F_l(w_l,X^i_1)\right)E\left(\mathcal{E}^i_1\right)=0.
\]
Therefore:
\[
\sup_{w_l\in W_l}\left|M^i_l(g,w_l)_m-M_l(g,w_l)\right|\rightarrow ^{a.s.}_{m\rightarrow \infty }0,
\]
which means that $P(B^i_l(g))=1$.

As $C(Z,\R{})$ is separable, there is a sequence $(h_t)_{t\in\N{}}$ dense in
$C(Z,\R{})$ (for the uniform norm). Let us denote
$A^i_l=\Omega_l\cap\bigcap_{t\in N{}}B^i_l(h_t)$. $A^i_l$ is measurable and
$P(A^i_l)=1$. Obviously, 
$B^i_l\cap\Omega_l\subset A^i_l$. Let us now show that
$B^i_l\cap\Omega_l=A^i_l$.  

Let $\omega\in A^i_l$. As $\omega\in \Omega_l$,
$\frac{1}{m}\sum_{j=1}^md_l(X^i_j(\omega))$ is a convergent sequence and
is therefore bounded, so there is $\gamma^i_l(\omega)>1$ such that for all $m$,
$\left|\frac{1}{m}\sum_{j=1}^md_l(X^i_j(\omega))\right|<\gamma^i_l(\omega)$.
Moreover, we can choose $\gamma^i_l(\omega)$ such that $\gamma^i_l(\omega)>E(d_l(X^i_1))$.

Let $g\in C(Z,\R{})$. For any $\epsilon>0$, there if
$t\in\N{}$ such that $\rho_Z(g,h_t)<\frac{\epsilon}{3\gamma^i_l(\omega)}$. This
obviously imply for all $w_l\in W_l$ and for all $m$ both
$|M^i_l(w_l,g)(\omega)_m-M^i_l(w_l,h_t)(\omega)_m|<\frac{\epsilon}{3}$  and
$|M_l(w_l,g)-M_l(w_l,h_t)|<\frac{\epsilon}{3}$. As $\omega\in A^i_l$,
$M^i_l(w_l,h_t)(\omega)_m$ converges to $M_l(w_l,h_t)$ uniformly on
$W_l$. Therefore there is $M$ 
such that $m>M$ implies
$\sup_{w_l\in
  W_l}|M^i_l(w_l,h_t)(\omega)_m-M_l(w_l,h_t)|<\frac{\epsilon}{3}$. Then $m>M$ 
implies $\sup_{w_l\in
  W_l}|M^i_l(w_l,g)(\omega)_m-M_l(w_l,g)|<\epsilon$. As this is true for
any $\epsilon$, we conclude that $M^i_l(w_l,g)(\omega)_m$ converges uniformly
on $W_l$ to
$M_l(w_l,g)$, and therefore that $\omega\in B^i_l(g)\cap\Omega_l$. As this is
true for all $g$, $\omega\in B^i_l\cap\Omega_l$. Therefore,
$B^i_l\cap\Omega_l=A^i_l$, which gives the conclusion of the lemma.
\end{pf*}

\textbf{part 3}\nopagebreak[4]

Let us now apply corollary \ref{theoremUniformSLLN} to
$\widehat{\lambda}_n(w)$, more precisely to the function from
$W\times(C(Z,\R{})\times \R{o})$ to \R{} define by:
\begin{multline*}
 k(w,g,t)=\\
c\left(t,U\left(w_0, \int F_1\left( w_1,x\right)
    g(x) \td\mu(x),\ldots ,\int F_k\left( w_k,x\right)
    g(x) \td\mu(x)
  \right) \right).
\end{multline*}
This is possible according to the following reasons:
\begin{itemize}
\item $W$ is compact
\item $k$ is continuous on $ w$ for each $(g,t) $, according to
  hypotheses $H_c$ and because, as a continuous function defined on a compact
  set, $ g$ belongs to $L^q(\mu)$.  Indeed, $w_l\mapsto \int F_l\left(
    w_l,x\right) g(x) \td\mu(x) $ is continuous for each $ g $: as $F_l$
  is continuous on $w$ for each $x$, the function $F_l\left(
    w'_l,.\right)g(.)$ converges punctually to $F_l\left( w_l,.\right)g(.)$ when
  $w'_l$ converges to $w_l$. Moreover, $\left|F_l\left( w,.\right)g(.)\right|$
  is dominated on $W_l$ by $d_l(.)|g(.)|$, which is integrable (by
  hypothesis) .  Thanks to dominated convergence theorem, this obviously
  implies the continuity of $w_l\mapsto \int F_l\left( w_l,x\right) g(x)
  \td\mu(x)$.
\item $ k$ is measurable with respect $ (g,t) $ for each $ w $.
This is a direct consequence of hypotheses $H_c$ and of the
fact that $g\mapsto \int F_l\left( w_l,x\right) g(x) \td\mu(x)$
is continuous for each $ w_l $
\item hypothesis $H_d$ implies that $k(w,g,t)\leq c_{max}(t)$ for all $w$, $g$
  and $t$, with $E(c_{max}(T_1))<\infty$
\item $(G^i,T^i)_{i\in\N{}}$ is i.i.d.
\end{itemize}
According to corollary \ref{theoremUniformSLLN}, we therefore have 
\[
\sup _{w\in W}\left| \frac{1}{n}
\sum ^{n}_{i=1}k(w,G^i,T^i)-E(k(w,G^1,T^1))\right| 
\rightarrow ^{a.s.}_{n\rightarrow \infty }0,
\]
that is
\begin{equation}\label{eqUnivConvLevelTwo}
\sup _{w\in W}\left| \widehat{\lambda}_n(w)-\lambda(w)\right| 
\rightarrow ^{a.s.}_{n\rightarrow \infty }0.
\end{equation}
Let us call $C$ the set of probability 1 for which this uniform convergence
occurs. Let us now consider $D=C\cap\bigcap_{i\in\N{}\ l\in\N{}}
(B^i_l\cap\Omega_l)$. According to lemma \ref{lemma:unif:integ}, $P(D)=1$. Let
$\omega$ be an arbitrary element of $D$ and denote for simplicity
$g^i=G^i(\omega)$ and $t^i=T^i(\omega)$. Let $ \epsilon >0 $ be an
arbitrary real number. According to equation \ref{eqUnivConvLevelTwo}, there
is $ N $ such that for each $ n\geq N $,
\begin{equation}\label{eqFirstLastStep}
\sup _{w\in W}\left|\frac{1}{n}\sum ^{n}_{i=1}k(w,g^i,t^i) -\lambda(w)\right|
<\frac{\epsilon }{2}.
\end{equation}
We handle here the case where $c$ is not a distance on $\R{o}$ but simply a
continuous positive function. As $U$ is bounded and uniformly continuous, the
function $l(t,w_0,u)=c(t,U(w_0,u))$ from $\R{o}\times W_0\times \R{k}$ is
uniformly continuous with respect to $(w_0,u)$. That is, for each $t^i$, there
is $\eta_i>0$ such that for each $w_0$ and 
$(u,u')\in{\R{k}\times\R{k}}$, $\|u-u'\|<\eta\Rightarrow
\|l(t^i,w_0,u)-l(t^i,w_0,u')\|<\frac{\epsilon}{2}$. As
$\omega\in \bigcap_{i\in\N{}\ l\in\N{}}
(B^i_l\cap\Omega_l)$, for each $i$, there is $S^i$ such that $ m^i\geq S^i $
implies for all $l$ 
\[
\sup_{w_l\in
  W_l}|M^i_l(w_l,g^i)(\omega)_{m^i}-M_l(w_l,g^i)|<\eta_i.
\]
Let us call $S_n=\sup_{i\leq n}S^i$. Then for $m\geq S_n$, for all $w$ and
for all $i\leq n$
\begin{multline*}
\bigl\|c\left(t^i,U\left(w_0,M^i_1(w_1,g^i)(\omega)_{m},\ldots,M^i_k(w_k,g^i)(\omega)_{m}\right)\right)\\
-c\left(t^i,U\left(w_0,M_1(w_1,g^i),\ldots,M_k(w_k,g^i)\right)\right)\bigr\|<\frac{\epsilon }{2},
\end{multline*}
that is for all $w\in W$
\begin{multline*}
\Biggl|\frac{1}{n}\sum
  ^{n}_{i=1}c\left(t^i,U\left(w_0,M^i_1(w_1,g^i)(\omega)_{m},\ldots,M^i_k(w_k,g^i)(\omega)_{m}\right)\right)\\
-\frac{1}{n}\sum ^{n}_{i=1}k(w,g^i,t^i)\Biggr|<\frac{\epsilon }{2}.
\end{multline*}
Combined with equation \ref{eqFirstLastStep}, this gives that for $n\geq N$
and $m\geq M(n)$:
\[
\sup_{w\in W}\left|\lambda _{n}^{m}(w)-\lambda(w) \right|<\epsilon.
\]
Therefore for almost all $\omega$ (i.e., for $\omega\in D$), we have:
\begin{equation}\label{equationUSLLN}
\lim _{n\rightarrow \infty }\lim _{m\rightarrow \infty }\sup
_{w\in W}\left| \lambda_{n}^{m}(w)-\lambda (w)\right| =0.
\end{equation}

\textbf{part 4} 

The final conclusion of the theorem is obtained exactly as in
\citet{White89Review}. We use the following lemma:
\begin{lem}\label{lemmaConvInf}
  Let $W$ be a compact set (considered with the metric $d$) and
  $(f_i^j)_{i\in\N{}\ j\in\N{}}$ a sequence of sequences of real valued
  continuous functions that converges uniformly to a continuous function $f$,
  that is $\lim _{n\rightarrow \infty }\lim _{m\rightarrow \infty
    }\rho_W(f_n^m,f)=0$.  Let us call $W^*$ the set of minimizers of $f$ and
  let $w_i^j$ be a minimizer of $f_i^j$. Then
\[
\lim _{n\rightarrow \infty }\lim _{m\rightarrow \infty }d(w_n^m,W^*)=0.
\]
\end{lem}
\begin{pf*}{Proof of lemma \ref{lemmaConvInf}}
First of all, it is clear that we just have to prove that the set of
accumulation points of $(w_i^j)_{i\in\N{}\ j\in\N{}}$, $Acc$, is included into
$W^*$. Indeed assume that both $Acc\subset W^*$ and that the conclusion of the
theorem does not hold. This implies that there is an infinite subsequence of
$(w_i^j)_{i\in\N{}\ j\in\N{}}$ which distance to $W^*$ remains above a fixed
positive number. As $W$ is compact this subsequence has at least one
accumulation point which cannot belong to $W^*$. As this accumulation point is
also an accumulation point of the full sequence, this contradicts our main
hypothesis.

Let us now consider $w^0$ an accumulation point of the sequence. Strictly
speaking, $w^0$ is the limit of a subsequence of the main sequence, but to
simplify the proof, we assume that $w^0=\lim _{n\rightarrow \infty }\lim
_{m\rightarrow \infty }w^m_n$. 

Let $\epsilon>0$. $f$ is uniformly continuous on $W$ and therefore there is $
\eta $ such that $ |w'-w|<\eta $ implies $ |f (w)-f(w')|<\epsilon $. By
uniform convergence, there is $N$ such that for each $n>N$, there is $M_n$
such that $m>M_n$ implies for all $w\in W$, $\left|f
  _{n}^{m}(w)-f(w)\right|<\epsilon$. Moreover, we can choose $N$ and $M_n$
such that $n>N$ and $m>M_n$ imply
$|w^m_n-w^0|<\eta$. Therefore, $n>N$ and $m>M_n$ imply
$|f_n^m(w^m_n)-f(w^0)|<2\epsilon$.

As $w^m_n$ is a minimizer of $f_n^m$, for all $w$, $f_n^m(w^m_n)-f_n^m(w)\leq
0$, which implies (by uniform convergence), $f_n^m(w^m_n)-f(w)\leq
\epsilon$. Therefore, $f(w^0)-f(w)\leq 3\epsilon$. As this is true for all
$\epsilon$, we conclude that $f(w^0)-f(w)\leq 0$ for all $w$ and therefore
that $w^0\in W^*$. Therefore $Acc\subset W^*$.
\end{pf*}
The conclusion of the theorem is obtained by applying lemma \ref{lemmaConvInf}
to all $\omega\in D$. For such a $\omega$, the uniform convergence of
$\lambda_{n}^{m}$ to $\lambda$ translates into the convergence of any
minimizer of $\lambda_{n}^{m}$ to the set of minimizers of $\lambda$. 
\end{pf*}
\newpage
\bibliographystyle{elsart-harv}
\bibliography{FunctionalMLP}

\end{document}